%% file: main.tex
\documentclass{article}
\usepackage{naturetex}
\usepackage{xurl}
\usepackage{xcolor}
\usepackage{amsfonts}
\usepackage{booktabs}

\begin{document}

\maketitle
{\setstretch{1.0}
	\section*{Abstract}
	    The development of powerful natural language models have increased the ability to learn meaningful representations of protein sequences. In addition, advances in high-throughput mutagenesis, directed evolution, and next-generation sequencing have allowed for the accumulation of large amounts of labeled fitness data. Leveraging these two trends, we introduce Regularized Latent Space Optimization (ReLSO), a deep transformer-based autoencoder which features a highly structured latent space that is trained to jointly generate sequences as well as predict fitness. Through regularized prediction heads, ReLSO introduces a powerful protein sequence encoder and novel approach for efficient fitness landscape traversal. Using ReLSO, we explicitly model the  sequence-function landscape of large labeled datasets and generate new molecules by optimizing within the latent space using gradient-based methods. We evaluate this approach on several publicly-available protein datasets, including variant sets of anti-ranibizumab and GFP. We observe a greater sequence optimization efficiency (increase in fitness per optimization step) by ReLSO compared to other approaches, where ReLSO more robustly generates high-fitness sequences. Furthermore, the attention-based relationships learned by the jointly-trained ReLSO models provides a potential avenue towards sequence-level fitness attribution information.
}

\input{sections/intro}
\input{sections/results}
\input{sections/discussion}

\input{sections/methods}



\bibliography{bibliography}
\bibliographystyle{naturemag}

\newpage

\section*{End Notes}
\subsection*{Acknowledgements}
We thank Alexander Tong and other members of the Krishnaswamy Lab for their helpful suggestions and discussion on this project. 


\subsection*{Declaration of Interests}
The authors declare no competing interests.

\subsection*{Data Availability}
The dataset used in this study are available in their processed form at the project's GitHub repository (\url{https://github.com/KrishnaswamyLab/ReLSO-Guided-Generative-Protein-Design-using-Regularized-Transformers/tree/main/data}). Additionally, we include links to the original data sources in the README file of the repository.

\subsection*{Code Availability}
Implementation of the models and optimization algorithms are available at the project's GitHub repository (\url{https://github.com/KrishnaswamyLab/ReLSO-Guided-Generative-Protein-Design-using-Regularized-Transformers}).

\newpage
\input{sections/sup}
\end{document}

%% file: sections/intro.tex
\section{Introduction}
\label{sec: introduction}

The primary challenge in sequence-based protein design is the vast space of possible sequences. A small protein of 30 residues (average length in eukaryotes $\approx$ 472 \cite{tiessen2012mathematical}) translates into total search space of $10^{38}$ -- far beyond to reach of modern high-throughput screening technologies. This obstacle is further exacerbated by \emph{epistasis} -- higher-order interactions between amino acids at distant residues in the sequence -- which makes it difficult to predict the effect of small changes in the sequence on its properties \cite{starr2016epistasis}. Together, this motivates the need for approaches that can better leverage sequence-function relationships, often described using \emph{fitness landscapes} \cite{romero2009exploring}, to more efficiently generate protein sequences with desired properties. To address this problem, we propose a data-driven deep generative approach called \emph{Regularized Latent Space Optimization} (ReLSO). ReLSO leverages the greater abundance of labeled data, arising from recent improvements in library generation and phenotypic screening technologies, to learn a highly structured latent space of joint sequence and structure information. Further, we introduce novel regularizations to the latent space in ReLSO such that molecules can be optimized and redesigned directly in the latent space using gradient ascent on the fitness function.


Although the \emph{fitness}\footnote{We use the term \emph{fitness} in general to refer to some quantifiable level of functionality that an amino acid sequence possesses, e.g. binding affinity, fluorescence, catalysis, and stability.} of a protein is more directly a consequence of the folded, three-dimensional structure of the protein rather than strictly its amino acid sequence, it is often preferable to connect fitness directly to sequence since structural information may not always be available. Indeed, when generating a library of variants for therapeutic discovery or synthetic biology, either through a designed, combinatorial approach or by random mutagenesis, it is cost-prohibitive to solve for the structure of each of the typically $10^{3}$ to $10^{9}$ variants produced. \\

Here we observe that protein design is fundamentally a search problem in a complex and vast space of amino acid sequences. For most biologically relevant proteins, sequence length can range from few tens to several thousands of residues \cite{tiessen2012mathematical}. Since each position of a $N$-length sequence may contain one of $20$ possible amino acids, the resulting combinatorial space ($\approx 20^N$ sequences) is often too large to search exhaustively. Notably this problem arises with the consideration of just canonical amino acids, notwithstanding the growing number of noncanonical alternatives \cite{chen2020engineering}. A major consequence of the scale of this search space is that most publicly-available datasets, though high-throughout in their scale, capture only a small fraction of possible sequence space and thus the vast majority of possible variants are left unexplored.\\


To navigate the sequence space, an iterative search procedure called directed evolution \cite{arnold_design_1998} is often applied, where batches of randomized sequences are generated and screened for a function or property of interest. The best sequence or sequences are then carried over to the next round of library generation and selection. Effectively, this searches sequence space using a hill climbing approach and as a consequence, is susceptible to local maxima that may obscure the discovery of better sequences. Other approaches to protein design include structure-based design \cite{rohl_protein_2004,norn_protein_2021}, where ideal structures are chosen \textit{a priori} and the task is to fit a sequence to the design. Recently, several promising approaches have emerged incorporating deep learning into the design \cite{brookes2018design, brookes2019conditioning}, search \cite{Yang:2019ix, biswas2021low}, and optimization \cite{linder2020fast} of proteins. However, these methods are typically used for \textit{in-silico} screening by training a model to predict fitness scores directly from the input amino acid sequences. Recent approaches have also looked to reinforcement learning to optimize sequences \cite{angermueller2019model}. Although these approaches are valuable for reducing the experimental screening burden by proposing promising sequences, the challenge of navigating the sequence space remains unaddressed.\\


An alternative to working in the sequence space is to learn a low dimensional, semantically-rich representation of peptides and proteins. These \emph{latent representations} collectively form the \emph{latent space}, which is easier to navigate. With this approach, a therapeutic candidate can be optimized using its latent representation, in a procedure called \emph{latent space optimization}. \\

Here we propose \emph{Regularized Latent Space Optimization} (ReLSO), a deep transformer-based approach to protein design, which combines the powerful encoding ability of a transformer model with a bottleneck that produces information-rich, low dimensional latent representations. The latent space in ReLSO besides being low dimensional is regularized to be 1) smooth with respect to structure and fitness by way of fitness prediction from the latent space, 2) regularized to be continuous and interpolatable between training data points, and 3) pseudo-convex based on negative sampling outside the data. This highly designed latent space enables optimization directly in latent space using gradient ascent on the fitness and converges to an optima that can then be decoded back into the sequence space.\\ 

Key contributions of ReLSO include:

\begin{itemize}
    \item The novel use of a transformer-based encoder with an autoencoder-type bottleneck for rich and interpretable encodings of proteien sequences
    \item A latent space is organized by sequence-function relationships, which ameliorates difficulties in optimization due to combinatorial explosion.
    \item A convex latent space that is reshaped using norm-based negative sampling to induce a natural boundary and stopping criterion for gradient-based optimization. 
    \item An interpolation based regularization which enforces gradual changes in decoded sequence space when traversing though latent space. This allows for a more dense sampling of the underlying sequence manifold on which the training data lies.
    \item A gradient ascent algorithm for generating new sequences from the latent space
\end{itemize}

We evaluate ReLSO on several publicly-available protein datasets, including variant sets of anti-ranibizumab and GFP. We view this domain first through a protein representation learning perspective, where we compare popular representations of proteins. We observe that ReLSO learns a more organized, smoother representation relative to other approaches. Next we examine the optimization ability of ReLSO on several protein design tasks. Compared to other sequence-based optimization approaches, ReLSO shows greater optimization efficiency (increase in fitness per optimization step) using it's fitness-directed traversal of latent space. This optimization efficiency allows ReLSO to more robustly generates high-fitness sequences. Lastly, the attention-based relationships learned by the jointly-trained ReLSO models provides a potential avenue towards sequence-level fitness attribution information. 

%% file: sections/results.tex
\section{Results}
\label{sec: results}




\begin{figure}
    \centering
    \begin{tabular}{cc}
    \includegraphics[width=0.95\textwidth]{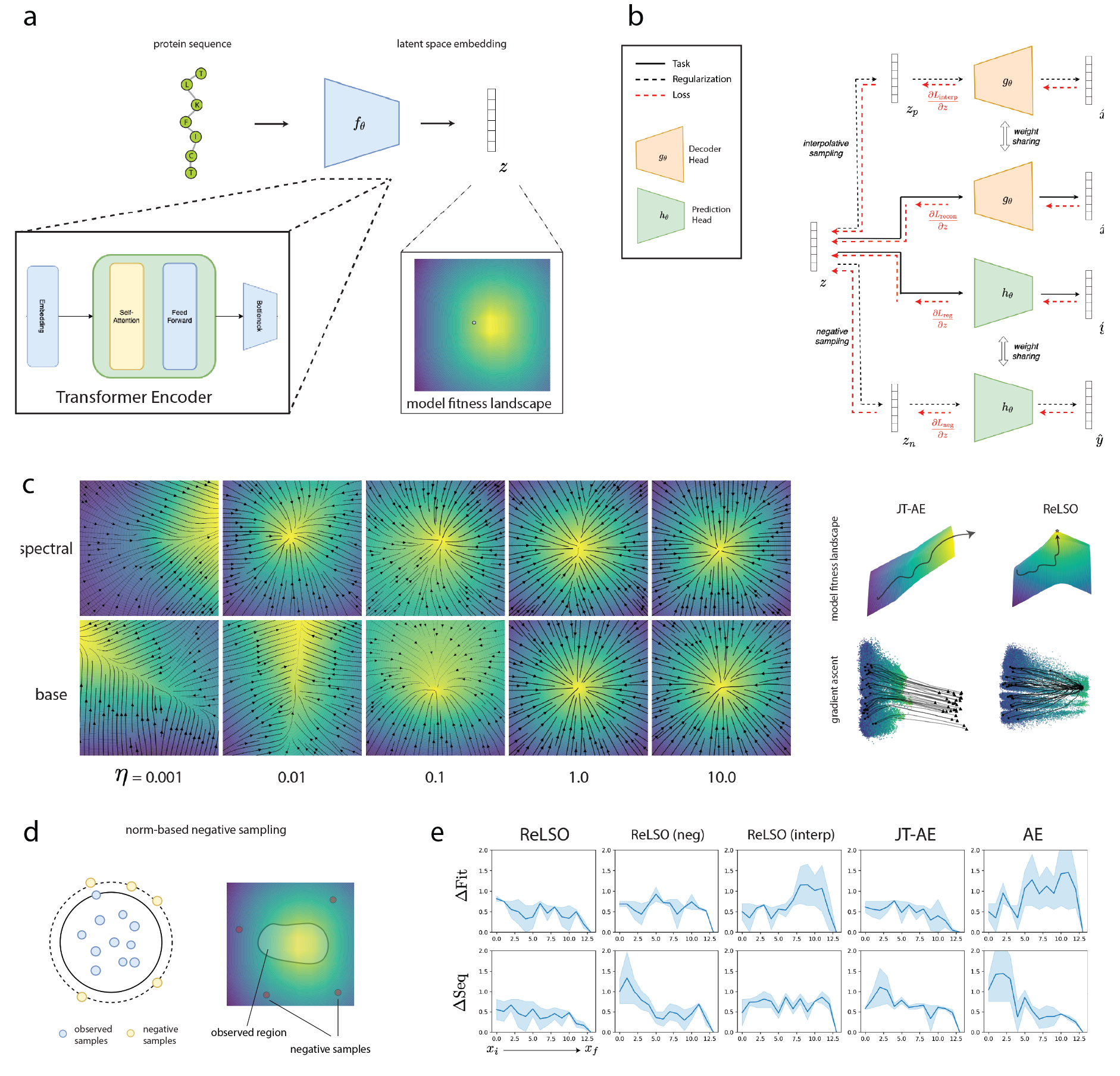}
    \end{tabular}
    \caption[textfont=small]{(A) Overall flow of information in ReLSO. Foundational to ReLSO is the idea that protein sequences can be mapped to a latent fitness landscape, which is maintains better properties than the traditional sequence-level fitness landscape. (B) To encourage a highly structure latent fitness landscapes, amenable to latent space optimization, ReLSO incorporates several regularizations (C) Negative samping regularization, weighted by hyperparameter $\eta$, restricts the overall shape of the fitness function to a psuedo-concave shape. An inherent weakness of a joint-training set-up is that a monotonic function is learned, which lacks any stopping criterion when used for latent space optimization, as in JT-AE. To address this, we reshape the fitness function such that the global maxima lies in/near the training data. (D) Negative sampling relies on a data augmentation strategy wherein artificial low-fitness points are generated (E) Smoothness along 100 sampled walks between pairs of distant points in latent space. Each step $x_i$ is along a nearest neighbor search to a final latent point $x_f$. $\Delta$Fit denotes L1 distance between each step's fitness and the final point's fitness. Similarly, $\Delta$Seq denotes Hamming distance of each step's sequence to the final step's sequence.}
    \label{fig: nmi-arch}
\end{figure}

\subsection{The ReLSO Architecture and Regularizations}
\label{sec: result-arch}

The ReLSO architecture is designed to jointly generate protein sequences as well as predict fitness from latent representations. The model is trained using a multi-task loss formulation which organizes the latent space by structure (input sequence) and function simultaneously, thus simplifying the task of finding sequences of high fitness from a search problem in a high-dimensional, discrete space to a much more amenable optimization problem in a low dimensional, continuous space. ReLSO leverages a transformer encoder to learn the mapping of sequences to latent space and utilizes gradient-based optimization to systematically and efficiently move through latent space towards regions of high-fitness. A norm-based negative sampling penalty is used to reshape the latent fitness landscape to be pseudo-convex. This has the dual benefit of further easing the optimization challenge as well as creating an implicit trust boundary. ReLSO makes innovative use of an interpolation regularization that enforces smoothness with respect to sequence, whereby small perturbations to a latent representation correspond to minor changes in the reconstructed sequence. This is especially relevant to protein design as it allows for a dense sampling of the latent space for diverse protein sequence generation while retaining properties of interest.\\

\subsubsection{Transformer-based Encoder with Dimensionality Reduction} 
\label{sec: result-arch-encoder}

ReLSO employs a transformer-based encoder to learn the mapping from a sequence, $x$, to its latent representation, $z$ (Figure \ref{fig: nmi-arch}A). While other encoding methods that rely on convolutional and recurrent architectures have demonstrated success in this domain \cite{Alley:2019fe, liu2020antibody}, we chose to use a transformer for several key reasons. First, the inductive bias of the transformer architecture matches the prevailing understanding that protein function is a consequence of pairwise interactions between residues (e.g. catalytic triads of proteases). Indeed the transformer architecture has shown promise in several tasks relying on protein sequence for prediction \cite{jumper2021highly, rao2019evaluating, rao2020transformer, rives2021biological}.Secondly, the transformer's attention-based encoding scheme provides for interpretability by analysis of learned attention weights \cite{vig2020bertology,rao2020transformer}. Finally, transformers have demonstrated advantages in representing long sequences, as a consequence of viewing the entire sequence during the forward pass, thus avoiding the limitations inherent to recurrent neural network based encoders \cite{hochreiter1998vanishing}.\\

The encoder network in ReLSO, $f_{\theta}$, transforms input proteins to a token-level representation where each amino acid in the sequence is replaced by a positional encoding of fixed length. This representation is then compressed to a coarse, sequence-level, representation using an attention-based pooling mechanism which computes the convex sum of positional encodings. This approach is preferred over mean or max pooling since it is able to weight the importance of positions in the sequence without incurring the computational cost of more complex recurrent-based pooling based strategies. Different than most transformer encoders, in RELSO the dimensionality of this sequence-level representation is further reduced using a fully-connected network (Figure \ref{fig: nmi-arch}A). This amounts to passing the sequence information through an information bottleneck, resulting in an informative and low-dimensional latent representation, $z$. Low dimensionality of the latent representation is important, since latent space grows exponentially with increasing dimension making latent space optimization more challenging.

\subsubsection{Jointly-trained Autoencoder (JT-AE)} 
\label{sec: result-arch-jointtraining}

ReLSO incorporates two vital factors in protein design: 1) sequence and 2) fitness information. By jointly training an autoencoder with a prediction network. The original autoencoder architecture, comprised of a encoder $f_{\theta}$ and decoder $g_{\theta}$, is supplemented with a network $h_{\theta}$ which is tasked with predicting fitness from the latent representation $z$. The resulting  objective function of this set-up takes the form,

$$ \mathcal{L} = ||g_{\theta}(f_{\theta}(x)) - x|| + || h_{\theta}(f_{\theta}(x)) - y||, $$

which includes the reconstruction loss and the fitness prediction (regression) loss. At each backpropagation step, the encoder is updated with gradients from both losses and is therefore directed to encode information about sequence and fitness in the latent representation, $z$ (Figure \ref{fig: nmi-arch}B). Indeed, when the dimension of $z$ is set to some low value, $d << N$, the encoder is forced to include only the most salient information about sequence and fitness and induces a connection between the two in $z$. This property was first exploited in the biomolecular domain \cite{gomez2018automatic}, where a jointly-trained variational autoencoder generated latent encodings organized by chemical properties. Here we leverage the same strategy to establish a strong correspondence between the protein sequence and its fitness, which is later utilized for generating novel sequences with desired fitness. Note that we refer the model architecture trained with the reconstruction and fitness prediction losses as \emph{JT-AE} (jointly trained autoencoder). ReLSO refers to the complete model, which includes negative sampling and interpolation losses, described next.

\subsubsection{Negative Sampling for pseudo-convexity of latent space} 
\label{sec: result-arch-negsampling}

A fundamental challenge in performing optimization in the latent space is that the optimization trajectory can stray far from the training data into regions where the prediction accuracy of the model deteriorates, producing untrustworthy results. Recent work has proposed techniques to define boundaries for model-based optimization by imposing constraints like a sequence mutation radius \cite{biswas2021low} or by relying on model-derived likelihood values \cite{brookes2019conditioning}. While mutation radii are straightforward to implement, the significant variability of fitness levels even within the immediate mutational neighborhood of a protein sequence, makes such global thresholds less than ideal. Additionally, a mutational radii constraint can be oversimplified as high mutation count may potentiate functional benefit to the fitness of a protein (e.g., the B.1.1.529 Omicron Spike protein). \\ 

The fitness prediction head of the JT-AE provides directional information for latent space optimization. However, it does not impose any stopping criterion nor any strong notion of a boundary or fitness optima. Furthermore, the addition of an auxiliary attribute prediction task, e.g. fitness prediction, to an autoencoder often results in unidirectional organization of the latent space by that attribute \cite{gomez2018automatic}. In such cases (Figure \ref{fig: nmi-arch}C), following the gradient produces an optimization trajectory that extends far outside the training manifold with no natural stopping point. This may ultimately result in generated protein sequences that are ‘unrealistic’ with respect to other biochemical properties necessary for proper folding such as stretches of homopolymers or amino acid sequence motifs known to be deleterious to stability/solubility in solution. 

In order to fully leverage the gradient signal provided by the fitness prediction head, $h_\theta$, we introduce a bias in learned fitness function, $\hat{\phi}_z$, towards regions in latent space near the training data. This is done using a data augmentation technique called \emph{norm-based negative sampling}. Each latent representation, $z$, obtained from the training data is complemented with a set of negative samples, $z_n$. These negative examples are produced by sampling high-norm regions of the latent space surrounding real latent points (Figure \ref{fig: nmi-arch}D). By assigning these artificial points, $z_n$, low fitness and including them in the fitness prediction loss, the learned fitness function, $\hat{\phi}_z$, is reshaped in a manner where there is a single fitness maxima located in or near the training data manifold. Using this regularization, an implicit trust region formed, thus providing a natural stopping criterion for latent space optimization. Next, we train ReLSO using a latent encoding dimension of two, $z \in \mathbb{R}^2 $ and visualize the resulting, learned fitness function by $h_\theta$. We observe that the negative sampling regularization induces a psuedo-concave shape wherein high fitness latent points reside in the center (Figure \ref{fig: nmi-arch}C). Increasing the strength of this regularization using a hyperparameter $\eta$ further enforces this organization.  We will refer to the JT-AE model augmented with this regularization as \emph{ReLSO (neg)}.\\

\subsubsection{Interpolative Sampling Penalty for Latent Space Continuity} 
\label{sec: result-arch-interpsampling}

To further improve the latent space of our jointly-trained for protein sequence optimization, we also introduce a penalty which enforces smoother interpolation in latent space with respect to sequence modifications. This is appealing for sequence-based protein design as we would like to be able more densely sample latent space for both analysis of latent space optimization trajectories as well as enrichment of the areas of sequence space around high fitness sequences.\\

We enforce gradual changes in the decoded sequence space during latent space traversal by the addition of an interpolation regularization term. For this term, we take a fraction of batch of latent points and compute a KNN graph using pairwise euclidean distances. A set of new latent points $z_p$ are then generated by interpolating between nearest neighbors. This new set of points $z_p$ are passed through the decoder network $g_{\theta}$ to produce a set of decoded sequences $\hat{x}_p$. We then penalize by the distance between two sequences in $\hat{x}$ and their interpolant $\hat{x}_p$. Formally, this penalty calculated element-wise by:

$$ \mathcal{L}_{interp} = \textrm{max}(0, \frac{ ||\hat{x}_1 - \hat{x}_i || + ||\hat{x}_2 - \hat{x}_i||}{2} - ||\hat{x}_1 - \hat{x}_2||)$$

where $\hat{x}_1$ and $\hat{x}_2$ are nearest neighbors in latent space and $\hat{x}_i$ is the decoded sequence of the interpolated latent point. We will refer to the JT-AE model augmented with only this regularization as \textit{ReLSO (interp)}. Finally the full model, with both negative sampling and interpolative sampling regularization, is referred to as \textit{ReLSO}.

\begin{figure}[!t]
    \centering
    \begin{tabular}{cc}
    \includegraphics[width=0.9\linewidth]{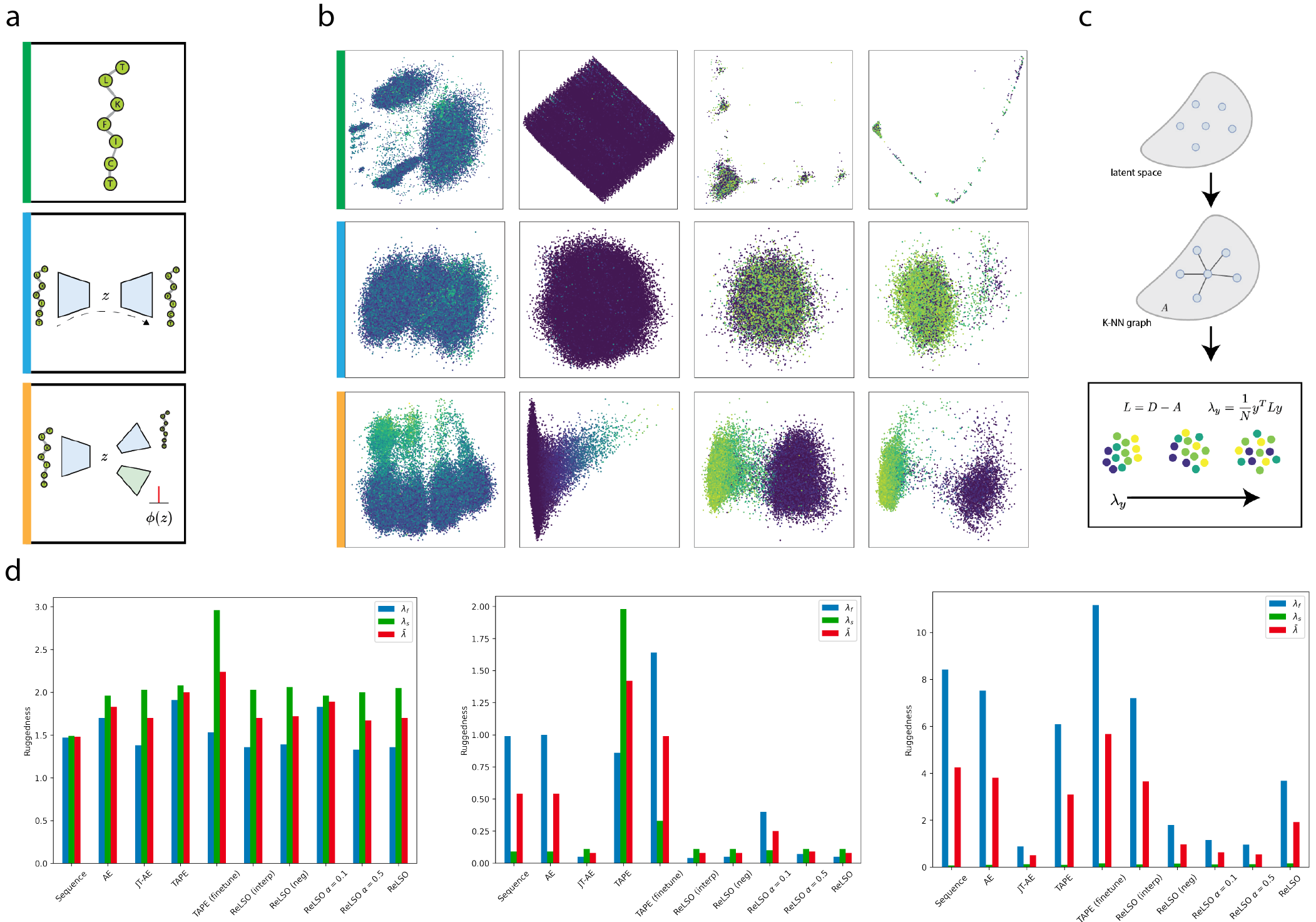}
    \end{tabular}
    \caption[textfont=small]{(A) We compare two main representation methods for proteins sequences, namely amino acid sequence identity (green edge) and latent embeddings from an unsupervised autoencoder (blue edge), to latent embeddings produced by a jointly-trained autoencoder (orange edge). (B) Grids of the three protein sequence representation approaches are visualized using PCA and colored by fitess. Across the four datasets (left to right: GIFFORD, GB1, GFP, TAPE) considered, joint-training produces a latent space organized by fitness and sequence information (C) Quantification of ruggedness is performed using a metric which measures neighborhood-level variation. (D) We compare the ruggedness of representations across several representation methods (See also Table \ref{tab:smooth-tab}).}
    \label{fig: nmi-smooth}
\end{figure}

\subsection{Latent Space Optimization and Sequence Generation Using ReLSO}

We leverage the highly-structured latent space of ReLSO to perform protein sequence optimization. With the negative sampling and interpolative sampling regularizations, the latent space of our model is organized by both fitness and sequence information(Figure \ref{fig: nmi-arch}A). Moreover, traversal in the latent space of ReLSO results in gradual changes in both sequence and fitness (Figure \ref{fig: nmi-arch}E). Combined these properties form a search space amenable to optimization approaches. \\

To optimize protein sequences, we use \textit{gradient-ascent} which allows for systematic and efficient modulation of fitness. First, a protein sequence $x$ is encoded by the encoder network, $f_\theta$ to produce a latent representation $z$. This process maps an input protein sequence to it's point in the model's latent fitness landscape. Next, the gradient with respect to the predicted fitness for the latent point, $\nabla_{z} h_{\theta}$, is calculated. The determined gradient provides directional information towards the latent fitness maxima and is used to update the latent point.

$$ z^{ (t+1) } \leftarrow z^{ (t) } + \epsilon \cdot \nabla_{z} h_{\theta} $$

This iterative process requires two hyperparameters, step size, $\epsilon$, and number of steps, $K$. At termination of the optimization loop, a final latent point $z_f$ is produced. This point in latent space is then decoded to it's corresponding sequence $x_f$ using the decoder module $g_{\theta}$. The reliance of a gradient signal over other, more stochastic approaches allows for a more robust approach to sequence optimization that is less sensitive to starting points. If greater sequence diversity is desired, we found that injecting noise into the update step can increase the variation in sequences produced (not shown). Overall, this process is referred to as \textit{latent space optimization} whereby protein sequences optimized in the latent space of a model rather than directly. A major benefit of using such an approach is the ability to learn a search space that ameliorates some of the challenges of optimizing protein sequences directly. In this work we learn better search space for protein sequence optimization by heavily regularizing an autoencoder-like model such that the latent space maintains favorable properties while still being generative.


.

\subsection{Ablation Study}
\label{sec: result-ablate}



ReLSO incorporates several components aimed at addressing challenges in protein sequence design. We examine the effect of removing major components through a set of ablation studies. These ablated versions of ReLSO include a standard autoencoder \textit{AE}, the base jointly-trained autoencoder without regularization \textit{JT-AE}, a ReLSO model without interpolation sampling \textit{RelSO (neg)} and a ReLSO model without negative sampling \textit{ReLSO (interp)}. We also include ReLSO with $\alpha \in [0.1,0.5]$ where $\alpha$ is a hyperparameter which weights the fitness prediction task and is by default set at $\alpha = 1$.\\

We first study the role of \emph{representation smoothness} in this domain where smoothness refers to the relationship between protein sequence representation and a property of interest. Properties considered here are sequence identity and fitness. We consider smoothness with respect to either of these properties as the difficulty of optimization is directly impacted by the smoothness of the search space. We first examine this qualitatively by viewing the three major representations of protein sequences used in protein sequence design - sequences of amino acids, latent encodings from a unsupervised model, and latent encodings from a jointly-trained model. These three representations are illustrated in Figure \ref{fig: nmi-smooth}A where latent encodings from the three representation approaches models are visualized using principle component analysis (PCA) and PHATE \cite{moon2017phate}. The smoothing effect that the joint-training approach generates can be observed. Furthermore, we observe that sequence distance can uninformative in cases such as that of GB1 where all sequences fall within a narrow mutational radius of one another.\\

We also examine representation smoothness quantitatively through the use of a metric derived from the graph signaling literature and previously used in the biomolecular domain to study thermodynamic landscapes \cite{castro2020uncovering}. In this approach pairwise distances between samples are computed to construct a KNN graph where edges of the graph reflect representation similarity. This graph can then be used to measure the variation of a signal along its edges such that the level of signal variation corresponds to the representation's ruggedness with respect to a signal of interest. The value of this metric is thus inversely proportional to the smoothness of a representation. These results are collected in Table \ref{tab:smooth-tab} where representations produced by ReLSO and its ablated forms are compared to the amino acid sequence representation. We also compare to the protein representations learned by the pre-trained TAPE transformer model from \cite{rao2019evaluating} which was trained on Pfam \cite{mistry_pfam_2021}.\\

As expected, the amino acid sequence representation possesses the highest smoothness with respect to sequence, however it's smoothness with respect to fitness is less than that of other approaches. This likely due to the often tangled relationship between mutational distance and fitness. Furthermore, models which have trained to predict fitness produce a smoother representation with respect to fitness than models trained solely on reconstruction.  With the ablations performed in ReLSO, it is observed that removal of the interpolation sampling regularization (ReLSO (neg)) reduces the smoothness with respect to sequence, This effect is also observed in Figure \ref{fig: nmi-arch}B where walks in latent space possess greater variation in sequence change when the interpolation sampling regularization is removed.\\

While the focus of this work is not to produce a new state-of-the-art fitness prediction model, it is useful to view differential performance of ReLSO and it's ablations. In our model set-up, we train on two tasks - a reconstruction task and a fitness prediction task. The first of requires that the model be able to generate sequences from the low-dimensional latent encoding. This allows the model to be generative, which is crucial in the domain of biomolecules as complete enumeration of sequence space is infeasible. We train the model using a Cross-Entropy loss across the possible amino acid identities of each position in the sequence. Here we report accuracy as well as perplexity to assess model performance in this task in Table \ref{tab:task-tab}. The second prediction head of this model is tasked with predicting the fitness value of input sequence. With this task, the model enforces the smoothness desideratum as well as learns a mapping from sequence to fitness. Due to the bottleneck introduced by the low-dimensional latent encoding $z$, the model produces a latent encoding there the coordinates of $z_i$ carry information about the fitness value of sequence $x_i$. We attempt to keep this mapping as simple as possible, and thus smooth, without losing fitness prediction performance by using a two-layer fully-connected network. We report these performance values in Table \ref{tab:task-tab}. It is observed that ReLSO's ability to predict fitness is not impacted by the inclusion of it's negative sampling based penalty. This is important as ReLSO's fitness prediction head guides the search process towards an optimally fit sequence. Moreover, the interpolative sampling regularization has a similar minor effect on reconstruction ability. We note that due the low diversity of GB1's protein sequences, it is difficult to meaningfully assess reconstruction ability. 


\subsection{Comparison to other protein sequence optimization strategies}
\label{sec: result-comparison}

\begin{figure}[t!]
    \centering
    \begin{tabular}{cc}
    \includegraphics[width=0.95\linewidth]{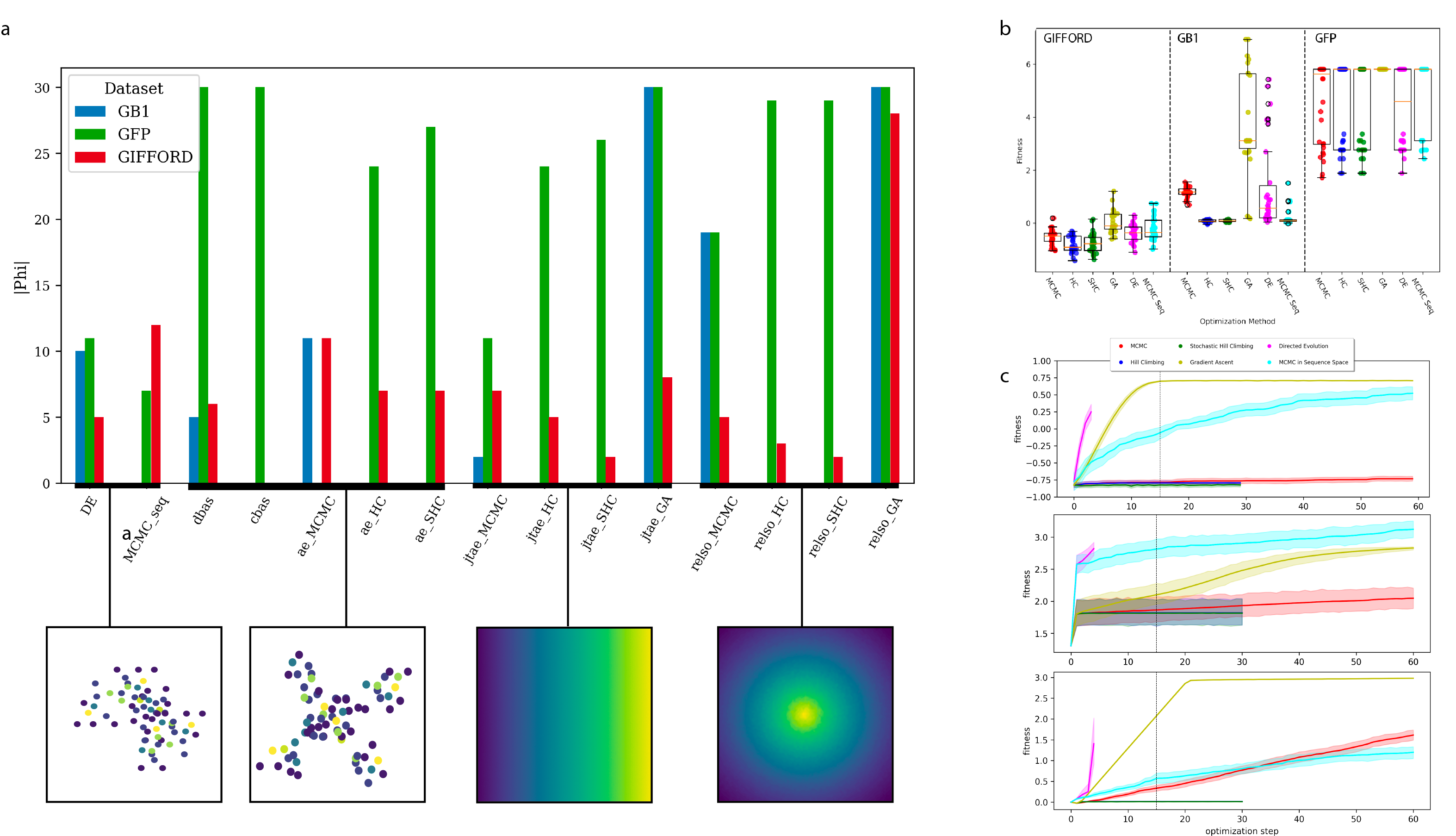}
    \end{tabular}
    \caption[textfont=small]{Comparison of methods for ML-based protein sequence optimization efficiency. (A) For each method, a batch of low-fitness sequences, sampled from the held-out set, were chosen as starting points. Ending optimized sequences that are predicted to be high-fitness, determined by a threshold, are included in a set $\Phi$. The optimization algorithms operate within a search space that is visualized below the plots. left to right: Sequence, AE, JTAE, ReLSO (B) Ending fitness values of all 30 seeds are plotted, partitioned by dataset. (C) The evolution of fitness values a over optimization step display a wide spread of efficiency. As some methods label multiple sequences during a single optimization step (e.g. HC, directed evolution) some lines do not each across the entire plot. To highlight gradient ascent's quick convergence to high-fitness, a vertical line is shown at optimization step in each plot}
    \label{fig: nmi-opt}
\end{figure}




In recent years, several protein sequence optimization methods have emerged that heavily rely on the use of a deep learning model. Some of these approaches use the model to perform an \textit{in-silico} screen of candidate sequences produced by iterative \cite{Yang:2019ix} or random \cite{biswas2021low} search. We view these methods as sequence-based search strategies as the generation of new sequence candidates occurs at the sequence-level. In contrast, methods such as \cite{brookes2018design, brookes2019conditioning} generate new sequences by sampling the latent space of a generative model. In this study, we seek to leverage the gradient information present in $h_\theta$ to search for more fit protein sequences. We observed that as a result of training to predict fitness from latent representation, the latent space organized by fitness, as show in Figure \ref{fig: nmi-smooth}B,D. To avoid several pathologies of optimization by gradient ascent in latent space, the regularizations included in ReLSO reshape this organization into a pseudo-concave shape (Figure \ref{fig: nmi-arch}D), which is eases the optimization challenge significantly.\\

As optimized sequences may possess hidden defects that present themselves in downstream analysis (e.g. immunogenicity of antibodies), it is often desired to produce several promising candidates at the end of the optimization stage. We replicate this scenario by collecting high-fitness sequences in a set $\Phi$ whereby inclusion is restricted to sequences which are predicted to have a fitness value above some threshold. We evaluate the considered optimization methods by the cardinality of each method's $\Phi$ (Figure \ref{fig: nmi-opt}A). Additional results on the composition of sequences found in each method's $\Phi$ are included in Table \ref{tab:opt-tab}, where we report a set of metrics which describe the quality and quantity of optimized sequences generated. For each optimization approach, we began with a set of 30 seed sequences drawn from the bottom 25th percentile of each dataset's held-out test split. Next, each optimization approach is allowed a \textit{in-silico} evaluation budget of 60 evaluations which corresponds to the number of sequences the method may observe and label. For methods like directed evolution which operate in a batchwise manner, the number of iterations is adjusted (Figure \ref{fig: nmi-opt}) to keep the total observed sequences constant across methods.\\

We find that ReLSO is able to produce a larger set of high-fitness sequences across the datasets with fewer optimization steps. This is observed in Figure \ref{fig: nmi-opt}C where gradient ascent (GA) is able to efficiently reach to regions of high-fitness latent space and consequently generate high-fitness sequences with a fraction of the evaluation budget expended. With the narrow fitness landscape of GFP, we observe that most methods are able to reach high-fluorescence sequences however in GIFFORD and GB1, the performance gap between gradient ascent and other methods is more pronounced. Furthermore,i n the case of GFP we observe that all optimized candidates of gradient ascent reside within the high-fluorescence mode of the dataset.

\begin{figure}[htb!]
    \centering
    \begin{tabular}{cc}
    \includegraphics[width=\linewidth]{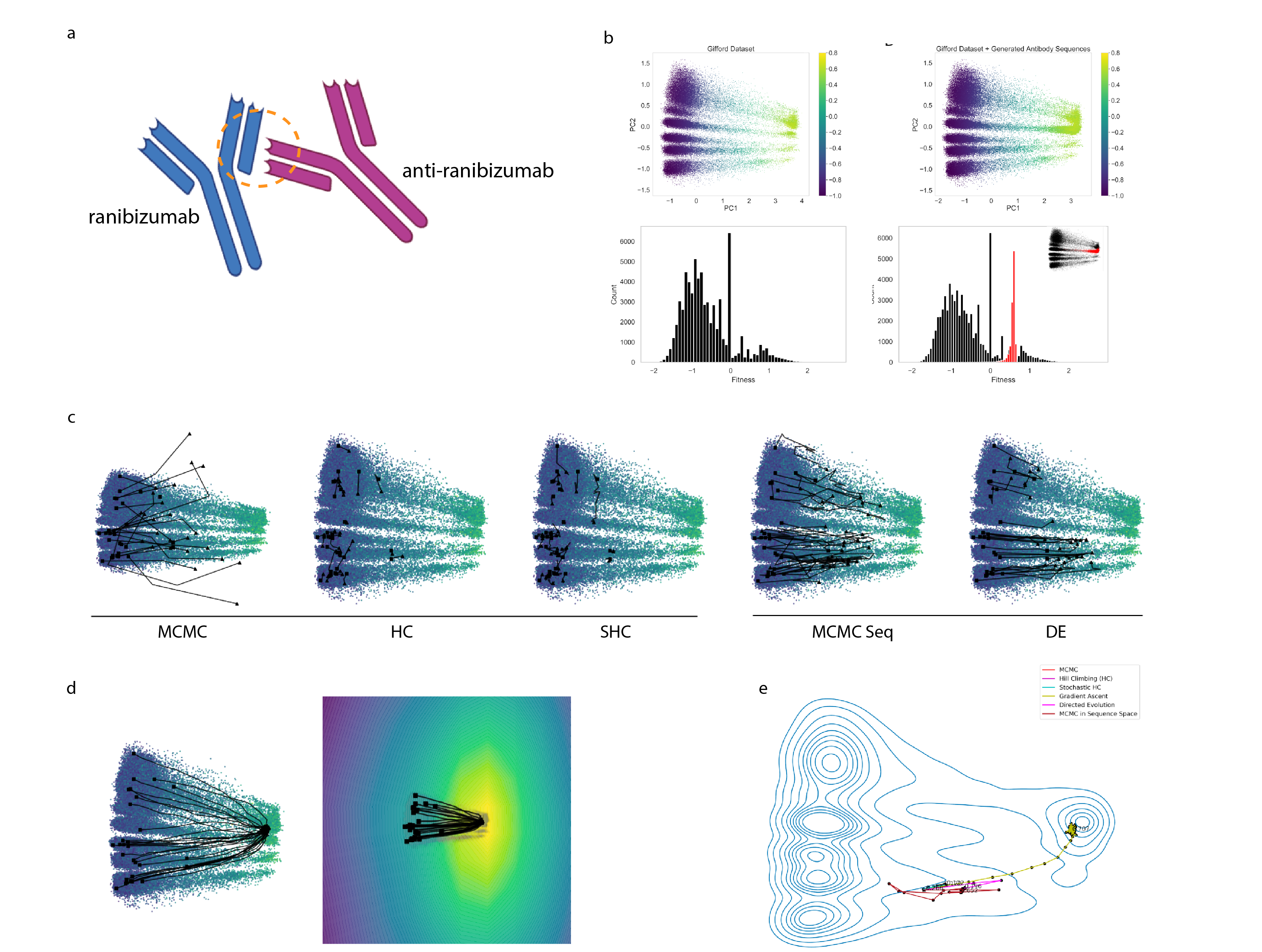}
    \end{tabular}
     \caption[textfont=small]{ (A) Antibody optimization of the  CDR3 region of an anti-ranibizumab antibody (B) Latent space visualization of high-fitness generated variants from the Gifford dataset. (A) Principal component analysis of the original training dataset model predicted embedding for each sequence. (B) Principal component addition of the 11,061 generated sequences onto the original latent space (yellow = high binding affinity/fitness, blue = low binding affinity/fitness). (C) Histogram distribution of fitness values for the original training dataset. Histogram distribution of fitness values of the predicted fitness values from the generated sequences overlapped onto the original dataset distribution for visualization (black = original, red = generated (C)Visualization of optimization paths in GIFFORD (A) Optimization trajectories of all 30 low-fitness seed sequences from the GIFFORD dataset from gradient-free methods (D) Optimization paths taken by gradient ascent converge to a high-fitness region of latent space. This is a result of the underlying regularized fitness function learned by ReLSO (E) For the optimization of a single seed, paths taken by gradient-free methods show a less-efficient progression to the data-driven global fitness maxima.}
    \label{fig: nmi-giff}
\end{figure}

\subsection{Optimization of Anti-Ranibizumab antibodies}
\label{sec: result-gifford}



To subsequently explore the ability of the model to generalize to similar but nonidentical sequences in the latent space, we generated $11,061$ sequences using the top sequences within the Gifford dataset (enrichment binding affinity $\approx 1.5$). This filtered to 43 high fitness ‘seed’ antibody sequences. Given the variability in physicochemical properties with respect to the potential of a single amino acid, we generated variants from each of these 43 sequences at one mutational amino acid distance away from these 43 starting sequences, which were of identical length and index position within the CDR3 region for consistency. Each of 20 possible amino acids was substituted at each position, without mutating more than one position for a given generated variant. This allowed exploration of the local neighborhood of the high fitness latent space while also localizing to a particular window of the CDR3 domain.\\

Upon interpolation of the high fitness sequences, each of 11,061 sequences was independently run through the trained model for both binding affinity and embedding predictions. The embedding of each sequence was obtained and plotted onto the latent space in the context of the entire training dataset. Interestingly, the generated sequences localized to a single high fitness region of the latent space. This indicated a denser sampling and fitness prediction of the local space using generated sequences that were not part of the initial dataset. Additionally, the generated sequence abundance appeared to be associated with increasing fitness as shown in Figure \ref{fig: nmi-giff}B, which was in line with the learned latent space embedding from the initial training data. These generative results suggest the potential implication of low mutational distances from regions of interest within the latent space of encoded sequences. Moreover, in the context of bio-therapeutic design, single amino acid manipulations may demonstrate significant differences in clinical efficacy. To this end, latent space optimization followed by a more dense, local generative sampling of high fitness variants may suggest an interesting avenue for sequence space exploration and bio-therapeutic optimization.\\


To subsequently explore the ability of the model to generalize to similar but nonidentical sequences in the latent space, we generated 11,061 sequences using the top sequences within the Gifford dataset (enrichment binding affinity $\approx 1.5$). This filtered to 43 high fitness ‘seed’ antibody sequences. Given the variability in physicochemical properties with respect to the potential of a single amino acid, we generated variants from each of these 43 sequences at one mutational amino acid distance away from these 43 starting sequences, which were of identical length and index position within the CDR3 region for consistency. Each of 20 possible amino acids was substituted at each position, without mutating more than one position for a given generated variant. This allowed exploration of the local neighborhood of the high fitness latent space while also localizing to a particular window of the CDR3 domain.\\

Upon interpolation of the high fitness sequences, each of 11,061 sequences was independently run through the trained model for both binding affinity and embedding predictions. The embedding of each sequence was obtained and plotted onto the latent space in the context of the entire training dataset. Interestingly, the generated sequences localized to a single high fitness region of the latent space. This indicated a denser sampling and fitness prediction of the local space using generated sequences that were not part of the initial dataset. Additionally, the generated sequence abundance appeared to be associated with increasing fitness as shown in Figure \ref{fig: nmi-giff}B, which was in line with the learned latent space embedding from the initial training data. These generative results suggest the potential implication of low mutational distances from regions of interest within the latent space of encoded sequences. Moreover, in the context of bio-therapeutic design, single amino acid manipulations may demonstrate significant differences in clinical efficacy. To this end, latent space optimization followed by a more dense, local generative sampling of high fitness variants may suggest an interesting avenue for sequence space exploration and bio-therapeutic optimization.


\begin{figure}[!htb]
    \centering
    \begin{tabular}{cc}
    \includegraphics[width=0.9\linewidth]{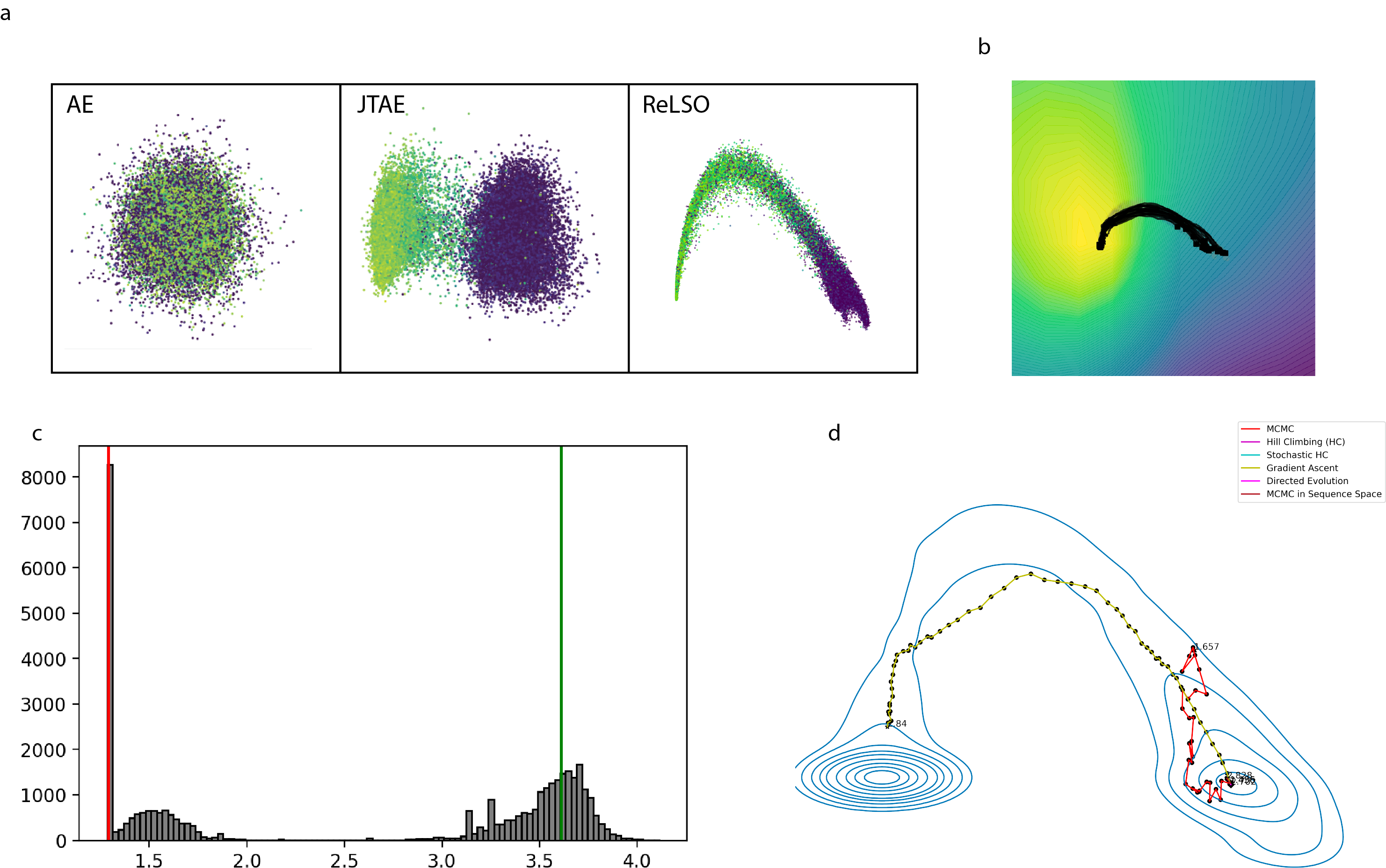}
    \end{tabular}
    \caption[textfont=small]{A) Visualization of the latent embeddings produced by AE, JTAE, and ReLSO model on the GFP dataset, colored by log fluorescence. (B) The learned fitness function extracted from $h_\theta$ network of ReLSO (C) Distribution of log fluorescence values presents a bimodal distribution with. a low and high fluorescence mode. (D) Optimization of a single, low-fluorescence seed sequence.}
    \label{fig: nmi-gfp}
\end{figure}

\subsection{GFP fitness landscape}
\label{sec: result-gfp}

In an effort to empirically describe epistasis in the fitness landscape of GFP, \cite{Sarkisyan:2016fy} performed random mutagenesis of the \textit{Aequorea victoria} GFP protein to produce 56,086 variants. The fluorescence of each variant was quantified, with multiple steps taken to ensure accurate estimates. The dataset produced from this study, which includes a mixture of variants with an average of 3.7 mutations per sequence, exhibited a narrow fitness landscape and a bimodal distribution of fluorescence values (Figure \ref{fig: nmi-gfp}C). The  sequence-fitness relationship in the dataset presented an opportunity to evaluate the ReLSO model on a protein known to have epistastic relationships between residues and a challenging optimization landscape. We observe the ruggedness of the fitness landscape in the sequence representation and latent representation generated by the AE model by PCA (Figure \ref{fig: nmi-smooth}A,B). However in the latent representations of JTAE and ReLSO the bimodal distribution of fitness in GFP fitness landscape is recapitulated (Figure \ref{fig: nmi-gfp}A). Investigation of the fitness function learned by the fitness prediction network of ReLSO, $h_\theta$, reveals that the model has learned a smoothly changing function from low to high fluorescence and a global optimum in the high-fluorescence mode (Figure \ref{fig: nmi-gfp}B).\\

With the latent space organized by fitness, as demonstrated by both visualization of the latent coordinates and the learned fitness function, we conducted \textit{in-silico} sequence optimization with the same setup used on the GIFFORD dataset. First we sampled seed sequences from the held-out test set and selected sequences which bottom quartile of observed fitness (log fluorescence $\leq$  1.3) as visualized by the red vertical line in (Figure Figure \ref{fig: nmi-gfp}C. The threshold for high-fitness and inclusion in $\Phi$ was set at the 95th percentile of log fluorescence (3.76) and is visualized with the green vertical line in Figure \ref{fig: nmi-gfp}C. Evaluation of optimization methods were then carried out and the results are presented in Figure \ref{fig: nmi-opt}A,B,C and Supp Table \ref{tab:opt-tab}. We observe a quick convergence to a high-fluorescence sequences by several methods, likely owing to the narrowness of the underlying fitness landscape \cite{Sarkisyan:2016fy}. However, knowledge of the fitness landscape still provides a reliable route for optimization as not all seed sequences were able to be successfully optimized by methods other than gradient ascent (Figure \ref{fig: nmi-gfp}D). For sequences predicted to be high-fluorescence by ReLSO, we observe a stabilizing change made to the sequence (avg ddG = -1.3kcal/mol) as determined by Dynamut2 \cite{rodrigues2021dynamut2}.

\begin{figure}[!t]
    \centering
    \begin{tabular}{cc}
    \includegraphics[width=0.9\linewidth]{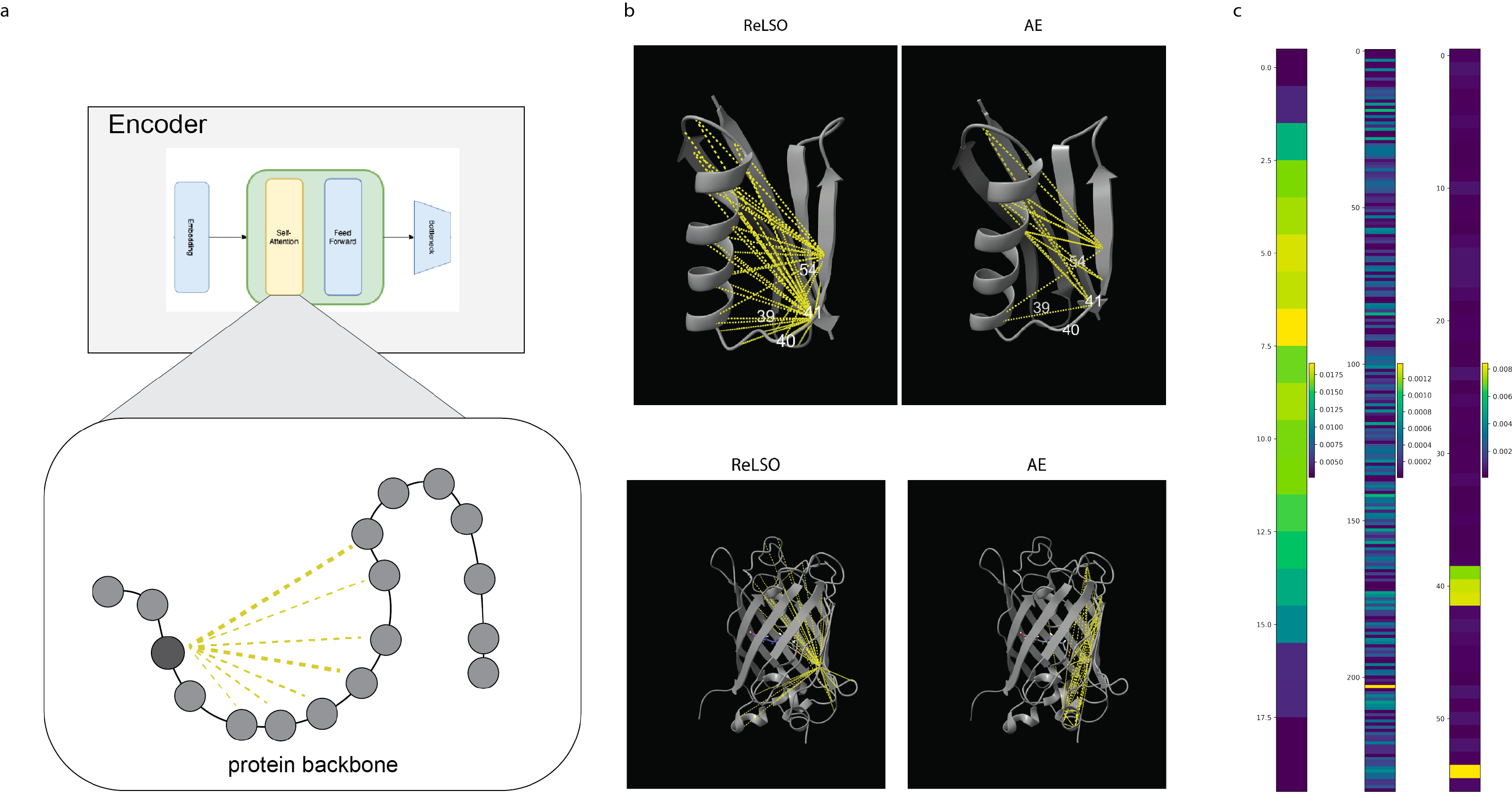}
    \end{tabular}
    \caption[textfont=small]{(A) Attention maps can be extracted from transformer-based encoder of ReLSO for interpretability analysis (B)Here we visualize learned pairwise relationships defined by the attention maps of the transformer encoder. We compare the attention maps of the ReLSO model to those of the AE model (Top) GB1 protein structure with overalayed attention and (Bottom) GFP protein structure with overalayed attention (C) To generate a sequence-level representation of the protein sequence, a attention-based pooling step is used in ReLSO. Importance of each position as determined by this pooling mechanism is show for GIFFORD, GFP, and GB1}
    \label{fig: nmi-attn}
\end{figure}


\subsection{Interpretability}
\label{sec: result-arch-interpret}

Encouraged by the success of other works of \cite{tamburini2020bertology} and \cite{rao2020transformer}, we examined the attention weightings of the trained ReLSO model for possible localized sequence-fitness attribution. We hypothesized that given the joint-training approach (Section \ref{sec: methods-setup}) and the observed organization by fitness in the latent embeddings (Figure \ref{fig: nmi-smooth}B), the learned attention information from the ReLSO model may also aid in elucidating fitness-sequence relationships. From the ReLSO model, we extract attention information from the attention maps of the transformer layers (Figure \ref{fig: nmi-attn}A) and the attention-based pooling layer. The former provides \textit{pairwise} attention information while the latter provides \textit{positional} attention information. To extract attention information, we randomly sample sequences from a dataset and pass them through the encoder and bottlneck modules of ReLSO. In each forward-pass, the attention maps of transformer model and the pooling weights of the bottlneck module are kept and aggregated. We then perform an averaging to reduce the level of noise. Additionally a thresholding step is performed on the averaged attention maps to increase sparsity such that only the most salient relationships remain. \\

As previous approaches have focused on transformers trained in an unsupervised or self-supervised manner, we compare the attention information between AE and ReLSO (Figure \ref{fig: nmi-attn}B). We observe several differences between the learned attention relationships of the ReLSO models and the AE models across the datasets. We view this divergence of attention relationships as a potential signal that the attention maps of transformers trained to predict fitness may also be probed to study sequence-fitness and structure-fitness relationships. Of note, the ReLSO attention maps for GFP appear to uniquely capture more distant epistatic interactions across the cylindrical pocket. These capture known epistasis in GFP \cite{Sarkisyan:2016fy} while also identifying potentially novel avenues for epistatic interrogation. Next, we are able to observe that the variable residues in the GB1 dataset (39-41, 54) are highly-weighted in the positional attention extracted from the ReLSO model, as shown in Figure \ref{fig: nmi-attn}C. As these 4 positions are the only changing residues across sequences in the GB1 dataset, we confirm their importance for the prediction tasks with their attention maps as expected. In addition, the embeddings of the amino acids recapitulate their underlying biochemical properties in their organization, shown in (Supp. Figure \ref{fig: aa-embeds}). This result has been shared in previous work as a signal that the model has learned biologically meaningful information in its parameters \cite{rives2021biological}. 

%% file: sections/discussion.tex
\section{Discussion}
\label{sec: discussion}

The ability to find better representations is vital to extracting insights from noisy, high-dimensional data within the fields of protein biology. Defined by their biochemical interactions, evolutionary selection pressures, and function-stability tradeoffs, biomolecules are an increasingly important domain for the application of deep learning. More specifically, the field of biotherapeutic development has recently shown significant benefits from the application of both linear and non-linear models. Some of the very impactful models in this space have been largely supervised, but more recent work has proven the usefulness of leveraging unsupervised learning to pre-train predictive models to identify protein sequences with an enhanced property of interest.\\

Here we took an alternative path to combining these two types of learning objectives by instead taking a multi-task learning approach. Through simultaneously optimizing for protein sequence generation and fitness level prediction, we explicitly enforce a latent space rich in information about both sequence and fitness information. Importantly, this fitness information may encompass a variety of different properties such as binding affinity and fluorescence, which is smoothly embedded in the latent space of our trained model.  We further add regularizations which reflect principles of protein engineering, reshaping the latent space in the process. Leveraging these regularizations and the architecture of the model, we show how gradient ascent optimization can deliver improvements in protein optimization when searching over protein sequence space.\\

The departure of this approach from other methods demonstrates a novel and promising avenue for improving our ability to design and optimize proteins. Furthermore, the reliance of this method solely on sequence information paired to a fitness value suggests ReLSO is likely translatable to other biomolecules such as DNA and RNA. The application to nucleic acids could have an interesting path towards reducing off-target effects in gene editing modalities such as CRISPR-Cas9. As such, with the growing prominence of biological therapeutics, this research direction has potential to deliver improvements in the development of improved therapeutics.\\

The intersection of machine learning and medicine as a whole has grown in recent years to suggest promise both in research as well as clinical practice. Diagnostic image data-rich fields such as pathology and radiology continue to pioneer towards optimized clinical management. Genomic data, and more recently, single-cell RNA sequencing data, have also pushed boundaries towards deciphering the complexity of genome regulation and target identification in hopes of better managing patient pathophysiology and disease. In the present study, we leveraged public protein fitness datasets to explore the boundaries of latent space optimized protein design. Moreover, while we explore several applications of our methodology and show robust generalized performance independent of a specific property, we envision an exciting application in biotherapeutic optimization. Specifically, we see an interesting avenue in tuning protein protein binding affinity to increase selectivity towards a certain target or isoform, but against others in an effort to mitigate off-target toxicity. Therapeutically modifying characterized diseased cell pathways using this methodology may have great potential in fields such as oncology, neurology, and infectious disease amongst others. Ultimately, the growing production of biotherapeutic fitness data, latent space optimization of therapeutic proteins may eventually improve the therapeutic management of patient disease as one exciting application.

%% file: sections/methods.tex
\section{Methods}
\label{sec: methodology}

\subsection{Model architecture and training details}
\label{sec: methods-setup}

ReLSO can be understood as being comprised of four main modules (\textit{Encoder Module, Bottlneck Module, Sequence Prediction Module, Fitness Prediction Module}). \textit{Encoder Module}. The encoder module takes as input protein sequences, encoded as an array of token indices, and outputs an array of token embeddings. The encoder module is a transformer \cite{vaswani2017attention} with 10 layers and 4 heads/layer. We use a token embedding size of 300 and hidden size of 400 in this work. \textit{Bottleneck Module} Next, amino acid-level embeddings are passed through a bottleneck module made up of two fully-connected networks. The first network reduces the dimensionality of each token to the latent dimension, 30, whereas the second predicts a vector of weights summing to 1. The outputs of these two networks are then combined in a pooling step where a weighted sum is taken across the 30-dimensional embeddings to produce a single sequence-level representation, $z \in \mathbb{R}^{30}$. This representation is of focus in the manuscript and is referred to as a protein sequence's \textit{latent representation}. \textit{Sequence Prediction Module} To decode latent points to sequences, we use a deep convolutional network, comprised of 4 1-dimensional convolutional layers. ReLU activations and batchnorm layers are used between convolutional layers with the exception of the final layer. \textit{Fitness Prediction Module} The final module is a fitness prediction network which predicts fitness from points latent in latent space. To encourage gradual changes to fitness in latent space we use a 2-layer fully connected network regularized by a spectral norm penalty introduced in \cite{yoshida2017spectral}. As a result, the network is further encouraged to learn simpler organizations such as the psuedo-concave shape described in this work.

We train each model for 300,000 steps with a learning rate of 0.00002 on two 24GB TITAN RTX GPUs, using a batch size of 64. For the negative sampling and interpolative sampling regularizations, samples of 32 artificial latent points are used in each forward pass bringing to total effective batch size to 128.

\subsection{Background and problem setup}
\label{sec: methods-setup}

From a preliminary experimental screen of variants, a set of protein sequences $\mathcal{X}$, where each sequence is a ordered set of amino acids $x = (\sigma_1, \sigma_2,...,\sigma_{N-1}, \sigma_{N})$ composed from a finite alphabet of amino acids such that $\sigma_i \in. V$, $i \in [N]$ and their corresponding fitness values $y$, $y \in \mathbb{R}$ is produced. The final dataset $\mathcal{D}$ is then comprised of pairs $(x_i, y_i)$ of sequences and their observed fitness. From this data, we wish to find sequences $x^* \in S$ that possess a high degree of fitness, as measured by some threshold $ \{ y^{*} \geq y_{thresh} \mid y^{*} = \phi(x^*), x* \in \Phi \}$. We also desire solutions in $\Phi$ to be diverse and novel.  

\subsection{Joint Training}
\label{sec: jt}

We formulate our method by first starting at the traditional perspective of sequence-based protein design. While directed evolution has yielded various successes in a range of domains over the years, it is susceptible to the underlying topology of the fitness landscape. This may lead to the accumulation of only locally optimal sequences at the completion of the optimization process. Recent work has sought to overcome the screening burden of directed evolution by performing \textit{in-silico} evaluations of a candidate sequence's fitness. This approach is comprised of training a model $\hat{\phi}_{x}$ to approximate the "ground-truth" fitness landscape $\phi_{x}$ by minimizing an objective function of the form $\mathcal{L} = ||\hat{y} - y||$, where $\hat{y} = \hat{\phi}_{x}(x)$ and $y = \phi_{x}(x)$. Once the model has converged, it is then used to evaluate sequence candidates $\bar{x}$ in either using either an iterative modification or sampling approach. In either case, the local sequence space around the original sequence is explored using minor changes to x, $\Delta_{x}$.  However, the difficult to predict relationship between $\Delta_{x}$ and changes in fitness $\Delta_{y}$ maintains the challenging nature of optimizing within sequence space.\\

A more recent approach to sequence-based protein design is to train a deep learning model to learn a representation of protein sequences by pre-training the model on a large corpus of protein sequence data with an unsupervised task $|| g(f(x)) - x||$ where $f_{\theta}$ is an encoder and $g_{\theta}$ is a decoder. The end result of this pre-training is a trained encoder that has learned a function $z = f_{\theta}(x)$, where $z$ is understood to contain abstract and useful information about protein sequence composition. For the unsupervised model, it can be further stated that learned latent code approximates the manifold on which the training data lies, where higher density is placed on realistic sequences.\\

Next, a prediction model $h_{\theta}$ is trained on the latent representation $z$ to learn a fitness landscape $\hat{y} = \hat{\phi}_{z} = h_{\theta}(z)$ that will be used to evaluate new candidates $\bar{x}$. While this approach moves the optimization problem to a continuous setting, it suffers from a issue inherent in datasets from this domain. Often the datasets gathered from screening contain a small percentage of high-fitness sequences while the vast majority of sequences provide little to no measurable functionality. This leads to high-fitness latent encodings that are hidden within a dense point cloud of low-fitness latent encodings, leading to inefficiencies when searching latent space through small perturbations to $z$, $\Delta_{z}$. As a result, the optimization process will spend much of its time traveling through dense regions of low-fitness candidates. Again, a strong link between changes in sequence information, now encoded in $\Delta_{z}$ and changes it's associated fitness $\Delta_{y}$ has yet to be established.\\

Here we propose to connect these two important factors through the use of an autoencoder model trained jointly on the fitness prediction task, thereby combining the described two-step process into one. In this way, we add to the autoencoder model architecture, comprised of a encoder $f$ and decoder $g$, a network $h$ which is tasked with predicting fitness from the latent representation $z$. The final objective function of this set-up takes the form 

$$ \mathcal{L}_{task} = \gamma \mathcal{L}_{recon} + \alpha \mathcal{L}_{reg} $$

where $\mathcal{L}_{recon}$ represents the reconstruction task and $\mathcal{L}_{reg}$ represents the fitness prediction task. Additionally, $\gamma$ and $\alpha$ are scalar hyperparameters which weight their respective tasks. We refer to this model as a \textit{JT-AE}. 

$$ z^{ (t+1) } \leftarrow z^{ (t) } - \eta \cdot (\nabla_{z}\mathcal{L}_{recon} + \nabla_{z} \mathcal{L}_{reg}) $$

An important consequence of the joint training set-up is that the latent code is updated during each train step with gradient signals from both sequence and fitness information. The resulting $z$ encoding is thereby induced to resolve the two pieces of information. After model convergence, the latent space is endowed with a strong sequence-fitness association which we leverage here for latent space optimization. 

$$ \mathcal{L} = ||g(f(x)) - x|| + || h(f(x)) - y|| $$

Furthermore, one can observe that in each update step, the encoder receives gradients from both the reconstruction loss and fitness prediction loss and is therefore directed to encode information about sequence and fitness in $z$. Indeed, when the dimensionality of $z$ is set to some low value $d << N$ the latent encoded is forced to include only the most salient information about sequence and fitness and induces a greater connection between the two in $z$. Through the use of this training strategy, we strengthen the connection between $\Delta_{z}$ and $\Delta_{y}$ for downstream applications.

\subsection{Negative Sampling}
\label{sec: neg-sampling}

A pervasive challenge of optimizing within the latent space of deep learning models is moving far from the training data into regions where the model's predictive performance deteriorates or is otherwise untrustworthy. Recent work has proposed techniques to define boundaries for model-based optimization, such as through a sequence mutation radius \cite{biswas2021low} or by relying on model-derived likelihood values \cite{brookes2019conditioning}. In general, the gradient's produced by a supervised network do not readily provide a stopping criterion nor any strong notion of bounds in regards to the range of values the network predicts. This can be further shown by training a network to predict from a 2-dimensional latent representation and overlaying the gradient directions onto the latent space. A unidirectional organization by the predicted attribute is the likely outcome, as shown shown by \cite{gomez2018automatic} and Figure \ref{fig: nmi-arch}C.\\

In this work we propose a solution which addresses the challenge of moving away from training points in latent space by focusing on the function learned by the neural network. During training, the model intakes  data in a batchwise manner where the batch of $N$ points that are randomly sampled from the training data. As output of the encoder module of ReLSO, these datapoints are encoded in a low-dimensional latent space. To keep latent embeddings close to the origin, we include a norm-based penalty on the encoding. This then allows for the generation of negative samples by randomly samping high-norm points in latent space. We sample $M$ latent points with L2-norms greater than that of the largest L2-norm observed in the original $N$ points. A hyperparameter is used to scale the allowed difference between the max L2-norm of the training samples and the min L2-norm of the negative samples. The training samples and negative samples are then concatenated batch which pass through the fitness prediction network. In the calculation of the mean-squared regression loss, fitness predicted for the negative samples are compared to a preset value. In this work we set this value as the minimum observed fitness in the dataset. The fitness prediction loss term is now the following, 

$$ \mathcal{L} =  \frac{1}{N}\sum_{t=1}^{n} (y_i - \hat{y}_i)^2 + \frac{1}{M}\sum_{t=1}^{m} (y_{neg} - min(Y))^2 $$

While negative sampling effectively restricts the function learned by fitness prediction network $h_{\theta}$ to resemble a concave shape, the ability of neural networks to learn a wide variety of functions can still allow for complex, non-concave shaped solutions to persist. In the bottom rows of Figure \ref{fig: nmi-arch}C, we further encourage learning of a smooth fitness function using spectral norm regularized layers from \cite{yoshida2017spectral}, labeled as "spectral". We compare this to a un-regularized fully-connected layer, labeled as "base". 


\subsection{Smoothness Index}
\label{sec: smoothness_index}


As \textit{smoothness} within the latent space encodings of our model play a major role in our approach, we measure smoothness with a metric used in \cite{castro2020uncovering}. We construct a symmetric KNN graph from the latent codes $Z = {z_i, z_j,...}$ from a set of sequences such that $z_i$ is $z_j$ are connected by an edge if either $z_i$ is within the K-nearest neighbors of $z_j$ or conversely, ifs $z_j$ is within the K-nearest neighbors of $z_i$. By constructing our graphs in this way, we ensure our metric is scale-invariant. The KNN graph $A$ is then used to construct the combinatorial graph Laplacian operator $L = D - A$ from which we calculate our smoothness metric as $$ \lambda_{y} =  \frac{1}{N}\,\,y^{T}Ly $$ where $y$ is our signal of interest and $N$ corresponds to the number of datapoints used to construct the graph. The quadratic form of the graph Laplacian operator can be interpreted as taking the sum of squared differences along edges in the underlying graph such that the resulting sum is lower if the signal is smooth i.e. small differences between neighboring points.


\subsection{Datasets}
\label{sec: methods-datasets}

Quantitative readouts of fitness landscapes have remained elusive until novel breakthroughs in high-throughput molecular biology, such as directed evolution and deep mutational scanning. Broadly speaking, these methods aim to introduce mutations in the sequence (or a set of interesting positions) in a systematic (saturation mutagenesis) or random (directed evolution) manner.

\subsubsection{Gifford Dataset}
\label{sec: methods-dataset-gifford}

Enrichment data from directed evolution; in this experiment, \cite{liu2020antibody} pitted a vast library (10\textsuperscript{10} unique mutants) of an antibody against a single target. This library was then pitted through three consecutive rounds of selection, washing, and amplification. Next-gen sequencing was used between rounds to identify which sequences were enriched  Here, the fitness is the log ratio of sequence enrichment between rounds of selection (i.e., how well the sequence performed relative to other members of the library).

\subsubsection{GB1 Dataset}
\label{sec: methods-dataset-gb1}

\cite{wu2016adaptation} carried out a saturation mutagenesis study targeted four sites and generated all 20\textsuperscript{4} possible mutants to explore the local fitness landscape of GB1, an immunoglobulin-binding protein. This particular site is known to be an epistatic cluster. Fitness was measured by testing stability and binding affinity. 

\subsubsection{GFP Dataset}
\label{sec: methods-dataset-gfp}

\cite{Sarkisyan:2016fy} carried out random mutagenesis on a fluorescent protein (avGFP) to generate 51,175 unique protein coding sequences, with an averages of 3.7 mutations. Fitness was determined by measuring fluorescence of mutated constructs via a fluorescence-activated cell sorting (FACS) assay. 

\subsubsection{TAPE Dataset}
\label{sec: methods-dataset-tape}

In addition to the datasets we pulled from prior work, we used the TAPE benchmark datasets for fluorescence from \cite{rao2020transformer}. Note that we also kept the train/test/validation splits consistent so as to establish a fair comparison. The data here is the same as \cite{Sarkisyan:2016fy} but is simply split by sequence distance.

\subsection{Optimization}
\label{sec: methods-opt}

\subsubsection{Sequence space optimization}
\label{sec: methods-opt-seq}

We compare against two popular approaches to ML-based protein sequence optimization. These methods manipulate sequences directly and use a machine learning model to screen candidates, effectively treating  model inference as a substitute for wet-lab characterization. The first of these methods which we consider is \textit{in-silico} directed evolution approach, as described in \cite{Yang:2019ix}. Here a subset of residues in a protein sequence of interest are iteratively expanded and screened \textit{in-silico}. The best amino acid for each position is then held constant while the next position is optimized. The second algorithm we consider which operates in sequence space is a Metropolis-Hastings Markov chain Monte Carlo approach used in \cite{biswas2021low} where protein sequences are undergo random mutagenesis and mutations are accepted with according to a probabilistic acceptance step. We refer to this approach as \textit{MCMC Seq} and the directed evolution approach as \textit{DE}.

\subsubsection{Gradient-free optimization}

The first set of gradient-free algorithms employ a local search where a small perturbation in the latent encoding is added $z_{t+1} = z_{t} + \epsilon$, where $t$ is the step index and $z_{t+1}$ is accepted with a probability equal to $\texttt{min}(1, exp(\frac{y_{t+1} - y_{t}}{kT})$. The second group of gradient-free optimization algorithms use a nearest-neighbors search and either move in the direction of the most fit neighbor (hill climbing) or chooses uniformly from the set $\Phi = {z_j | h(z_j) = y_j > y_i}$ (stochastic hill climbing). Since we train the fitness prediction head directly from the latent encoding, we have access to the gradients of this network and can perform gradient ascent. We also examine the effect of cycling candidates back through the model as denoted in equation 2. We also examine two gradient-free methods which operate in sequence space. One such method is a form of \textit{in-silico} directed evolution where positions are independently and sequentially optimized. The second optimization approach mirrors that of the metropolis-hasting Monte Carlo search approach used in latent with the exception that the perturbation step is replaced with a mutation step. 

\subsubsection{Gradient-free latent space optimization}
\label{sec: methods-opt-latent-wograd}

Next we consider a set of optimization algorithms that operate in the latent space of generative models. These methods still treat the prediction network as a black-box, unable to access gradients, but manipulate sequences using their latent encodings. First, we pursue a simple hill-climbing algorithm (\textit{HC}) which takes a greedy search through latent space. We also test a stochastic variant of hill climbing, which should better avoid local minima. Here the algorithm samples $z_{t+1}$ uniformly from $ \{z \, | \, h_{\theta}(z+\epsilon) > h_\theta(z_t) , \epsilon \sim \mathcal{N}(\mu,\, k) \}$, where $k$ is a parameter. We refer to this variation of hill climbing as stochastic hill climbing (\textit{SHC}). Furthermore, we use a MCMC scheme similar to the previously described approach of \cite{biswas2021low}, however here we perform this optimization in latent space. We apply a small perturbation in the latent encoding $z_{t+1} = z_{t} + \epsilon$, where $t$ is the step index and $z_{t+1}$ is kept according to a probabilistic acceptance step. Lastly, we consider the approach of \cite{brookes2018design} and \cite{brookes2019conditioning} where an adaptive sampling procedure is done to generate improved sequences. We denote these approaches as \emph{DbAS} and \emph{CbAS}.

\subsubsection{Gradient-based latent space optimization}
\label{sec: methods-opt-latent-grad}

We examine the performance of a latent space gradient ascent optimization approach. Here we leverage the ability to extract gradient directions provided by the jointly-trained fitness prediction head of the model $h_{\theta}$. These directions allow for latent space traversal to areas of latent space associated with higher fitness sequences. We first examine this approach through a jointly-trained autoencoder without the aforementioned regularizations and denote this approach as \textit{JTAE - GA}. Next, we examine the performance of gradient ascent with using the interpolation sampling and negative sampling regularizations of ReLSO and refer to this approach as \textit{ReLSO - GA}. 

\subsubsection{CbAS and DbAS}
\label{sec: methods-opt-cbas-dbas}

In this work we also compare to the \textit{DbAS} and \textit{CbAS} methods introduced in \cite{brookes2018design} and \cite{brookes2019conditioning}, respectively. In this approach, the fitness prediction model is treated as a black-box "oracle" which maps from design space to a distribution over properties of interest. Similar to our approach, the authors of these methods consider the pathologies inherent to relying on deep learning models to optimize protein sequences. To address this, DbAS and CbAS use a model-based adaptive sampling technique which draws from the latent space of a generative model. While DbAS assumes an unbiased oracle, CbAS conditions the sampling procedure on the predictions of a set of oracle models. in this study, we use an implementation sourced from a publicly available \href{https://github.com/dhbrookes/CbAS}{github repo}. To ensure representative performance, we first evaluated DbAS and CbAs on the datasets using a grid search over several values of q ([0.5, 0.6, 0.7, 0.8]) and number of epochs using in training ([1,5,10,15]).


%% file: sections/sup.tex




\section*{Supplementary}

\begin{table}[!htb]
\centering
\resizebox{\textwidth}{!}{%
\begin{tabular}{@{}llllllllll@{}}
\toprule
\textbf{Search Space} & \textbf{Opt Algo} & \textbf{Max Fit.} & \textbf{Mean Fit.} & \textbf{Std Fit} & \textbf{$| \Phi |$} & \textbf{Novelty} & \textbf{Diversity} & \textbf{Ensgrad - Max} & \textbf{Ensgrad - Mean} \\ \midrule
S                     & DE \cite{Yang:2019ix}              & 0.30              & -0.37              & 0.33             & 11               & 1.00             & 0.08               & -0.16                  & -0.52                   \\
                      & MCMC Seq    \cite{biswas2021low}       & 0.75              & -0.18              & 0.45             & 13               & 1.00             & 0.12               & 0.12                   & -0.53                   \\ \midrule
L-AE                  & MCMC              & 0.00              & -0.39              & 0.23             & 11               & 1.00             & 0.08               & 0.23                   & -0.06                   \\
                      & HC                & 1.35              & -0.62              & 0.51             & 7                & 0.14             & 0.03               & 0.25                   & -0.64                   \\
                      & SHC               & 0.14              & -0.65              & 0.39             & 7                & 0.43             & 0.03               & -0.22                  & -0.64                   \\
L- JTAE               & MCMC              & 0.07              & -0.48              & 0.27             & 8                & 1.00             & 0.04               & -0.41                  & -0.28                   \\
                      & HC                & 0.24              & -0.61              & 0.35             & 5                & 1.00             & 0.01               & -0.46                  & -0.68                   \\
                      & SHC               & -0.13             & -0.70              & 0.28             & 1                & 0.00             & 0.00               & -0.02                  & -0.70                   \\
L - ReLSO             & MCMC              & 0.19              & -0.53              & 0.27             & 3                & 1.00             & 0.01               & 0.14                   & -0.39                   \\
                      & HC                & -0.30             & -0.81              & 0.30             & 0                & NA               & NA                 & -0.43                  & -0.70                   \\
                      & SHC               & 0.15              & -0.76              & 0.36             & 2                & 0.00             & 0.00               & -0.35                  & -0.69                   \\
L - VAE               & DBas   \cite{brookes2018design}           & 0.00              & -0.47              & 0.25             & 6                & 1.00             & 0.01               & -0.27                  & -0.58                   \\
                      & CBas     \cite{brookes2019conditioning}          & -0.30             & -0.62              & 0.24             & 0                & NA               & NA                 & -0.33                  & -0.62                   \\ \midrule
L - JTAE              & GA                & -0.55             & -0.56              & 0.06             & 0                & 0.00             & 0.00               & -0.02                  & -0.32                   \\
L - ReLSO             & GA                & 1.20              & 0.05               & 0.41             & 23               & 1.00             & 0.24               & 0.33                   & -0.01                   \\ \bottomrule
\end{tabular}%
}
\caption{Optimization of protein fitness. Here we evaluate the performance of three categories of optimization methods. The first three entries can be understood as optimization within latent space using local search approach. Latent space  Metropolis Hastings MCMC searches pockets of latent space whereas hill climbing (HC) and stochastic hill climbing (SHC) use the locations of nearby points for directions to move in. The second category made up of Metropolis Hastings MCMC (MCMC Seq) and directed evolution (DE) are optimization approaches which operate on the sequence directly. The final category is a latent space optimization approach which uses a gradient-guided search process, gradient ascent (GA). We observe generally greater performance with this third class of method in the proposed evaluation critera. }
\label{tab:opt-tab}
\end{table}


\begin{table}[]
\centering
\renewcommand{\arraystretch}{1.25}
\begin{tabular}{lrrrrrrrrr}
\hline
                   & \multicolumn{1}{l}{GIFFORD}     & \multicolumn{1}{l}{}            & \multicolumn{1}{l}{}                & \multicolumn{1}{l}{GB1}         & \multicolumn{1}{l}{}            & \multicolumn{1}{l}{}                & \multicolumn{1}{l}{GFP}         & \multicolumn{1}{l}{}            & \multicolumn{1}{l}{}                \\ \hline
                   & \multicolumn{1}{l}{$\lambda_f$} & \multicolumn{1}{l}{$\lambda_s$} & \multicolumn{1}{l}{$\hat{\lambda}$} & \multicolumn{1}{l}{$\lambda_f$} & \multicolumn{1}{l}{$\lambda_s$} & \multicolumn{1}{l}{$\hat{\lambda}$} & \multicolumn{1}{l}{$\lambda_f$} & \multicolumn{1}{l}{$\lambda_s$} & \multicolumn{1}{l}{$\hat{\lambda}$} \\ \cline{2-10} 
Sequence           & 1.47                            & 1.49                            & 1.48                                & 0.99                            & 0.09                            & 0.54                                & 8.42                            & 0.07                            & 4.25                                \\ \hline
AE   & 1.70                  & 1.96                      & 1.83                          & 1.00                   & 0.09                     & 0.54                      & 7.52                            & 0.10                            & 3.81                                \\
TAPE \cite{rao2019evaluating} & 1.91                            & 2.08                            & 2.00                                & 0.86                            & 1.98                            & 1.42                                & 6.09                            & 0.10                            & 3.09                                \\  \hline
JT-AE              & 1.38                            & 2.03                            & 1.70                                & 0.05                            & 0.11                            & 0.08                                & 0.88                            & 0.12                            & 0.50                                \\
TAPE + finetune    & 1.53                            & 2.96                            & 2.24                                & 1.64                            & 0.33                            & 0.99                                & 11.17           & 0.16            & 5.67              \\  \hline
ReLSO (interp)     & 1.36                            & 2.03                            & 1.70                                & 0.04                            & 0.11                            & 0.08                                & 7.20                            & 0.11                            & 3.65                                \\
ReLSO (neg)        & 1.39                            & 2.06                            & 1.72                                & 0.05                            & 0.11                            & 0.08                                & 1.80                            & 0.15                            & 0.97                                \\
ReLSO $\alpha=0.1$ & 1.83                            & 1.96                            & 1.89                                & 0.40                            & 0.10                            & 0.25                                & 1.15                            & 0.11                            & 0.63                                \\
ReLSO $\alpha=0.5$ & 1.33                            & 2.00                            & 1.67                                & 0.07                            & 0.11                            & 0.09                                & 0.96                            & 0.12                            & 0.54                                \\
ReLSO              & 1.36                            & 2.05                            & 1.70                                & 0.05                            & 0.11                            & 0.08                                & 3.68                            & 0.16                            & 1.92                                \\ \hline
\end{tabular}
\vspace{1em}
\caption{Quantification of latent space ruggedness, described in Section \ref{sec: result-ablate}. Ruggedness values with respect to fitness ($\lambda_{f}$) and sequence ($\lambda_{s}$). The average of those two values $\bar{\lambda} = (\lambda_{f} + \lambda_{s})/2 $ is also reported.}
\label{tab:smooth-tab}
\end{table}

\begin{table}[hbt!]
\renewcommand{\arraystretch}{1.25}
\centering
\resizebox{\textwidth}{!}{%
\begin{tabular}{@{}lrrrrrrrrrrrr@{}}
\toprule
                     & \multicolumn{1}{l}{\textbf{Gifford}} & \multicolumn{1}{l}{}         & \multicolumn{1}{l}{}       & \multicolumn{1}{l}{}                & \multicolumn{1}{l}{\textbf{GB1}} & \multicolumn{1}{l}{}         & \multicolumn{1}{l}{}       & \multicolumn{1}{l}{}                & \multicolumn{1}{l}{\textbf{GFP}} & \multicolumn{1}{l}{}         & \multicolumn{1}{l}{}       & \multicolumn{1}{l}{}                \\ \midrule
                     & \multicolumn{1}{l}{Task 1}           & \multicolumn{1}{l}{}         & \multicolumn{1}{l}{Task 2} & \multicolumn{1}{l}{}                & \multicolumn{1}{l}{Task 1}       & \multicolumn{1}{l}{}         & \multicolumn{1}{l}{Task 2} & \multicolumn{1}{l}{}                & \multicolumn{1}{l}{Task 1}       & \multicolumn{1}{l}{}         & \multicolumn{1}{l}{Task 2} & \multicolumn{1}{l}{}                \\ \cmidrule(l){2-13} 
                     & \multicolumn{1}{l}{Perplexity}       & \multicolumn{1}{l}{Accuracy} & \multicolumn{1}{l}{MSE}    & \multicolumn{1}{l}{Spearman $\rho$} & \multicolumn{1}{l}{Perplexity}   & \multicolumn{1}{l}{Accuracy} & \multicolumn{1}{l}{MSE}    & \multicolumn{1}{l}{Spearman $\rho$} & \multicolumn{1}{l}{Perplexity}   & \multicolumn{1}{l}{Accuracy} & \multicolumn{1}{l}{MSE}    & \multicolumn{1}{l}{Spearman $\rho$} \\
                     \cmidrule(l){2-13}

AE & 1.03                                 & 0.90                         & 0.88                       & -0.15                               & 1.00                             & 1.00                         & 0.17                       & 0.00                                & 1.00                             & 0.99                         & 6.74                       & 0.13                                \\
JT-AE                & 1.21                                 & 0.82                         & 0.22                       & 0.47                                & 1.00                             & 1.00                         & 0.01                       & 0.43                                & 1.04                             & 0.99                         & 0.18                       & 0.85                                \\
ReLSO (interp)       & 1.21                                 & 0.82                         & 0.22                       & 0.48                                & 1.00                             & 1.00                         & 0.01                       & 0.43                                & 1.03                             & 0.99                         & 0.13                       & 0.86                                \\
ReLSO (neg)          & 1.24                                 & 0.81                         & 0.29                       & 0.47                                & 1.00                             & 1.00                         & 0.02                       & 0.42                                & 1.09                             & 0.98                         & 0.22                       & 0.77                                \\

ReLSO $\alpha = 0.1$ & 1.02                                 & 0.91                         & 0.72                       & 0.35                                & 1.00                             & 1.00                         & 0.09                       & 0.53                                & 1.03                             & 0.99                         & 0.18                       & 0.84                                \\
ReLSO $\alpha = 0.5$ & 1.07                                 & 0.88                         & 0.34                       & 0.50                                & 1.00                             & 1.00                         & 0.02                       & 0.45                                & 1.04                             & 0.99                         & 0.12                       & 0.85                                \\ \cmidrule(l){2-13} 
ReLSO                & \textbf{1.17}                        & 0.84                         & 0.29                       & 0.48                                & 1.00                             & 1.00                         & 0.01                       & 0.44                                & 1.10                             & 0.98                         & 0.52                       & 0.70                                \\ \bottomrule
\end{tabular}%
}
\vspace{1em}

\caption{Task performance across the two tasks. Task 1 has the model predict sequence from latent space whereas Task 2 requires the model to predict fitness}
\label{tab:task-tab}
\end{table}

\begin{figure}[!h]
    \centering
    \begin{tabular}{cc}
    \includegraphics[width=\linewidth]{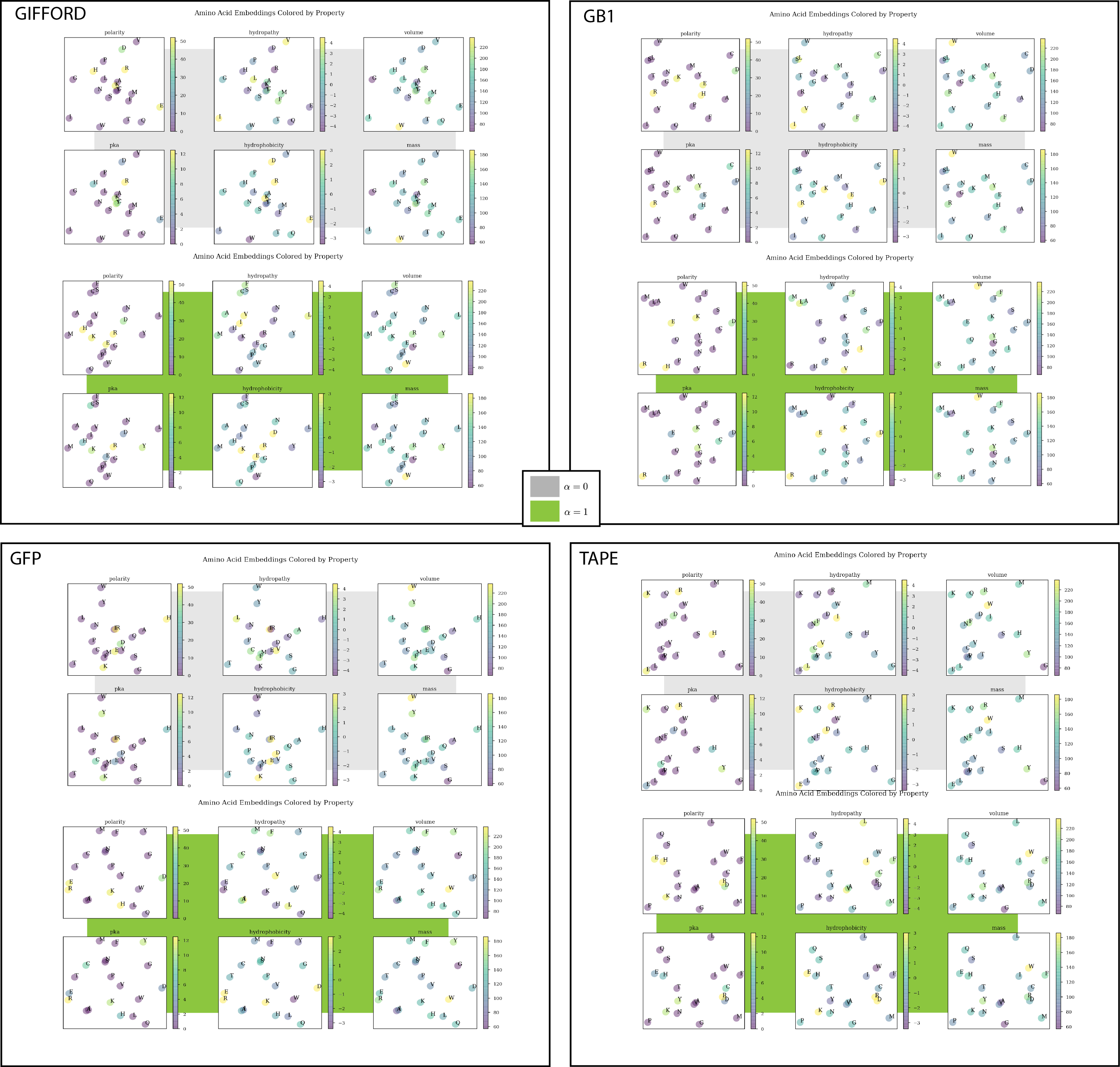}
    \end{tabular}
    \caption[textfont=small]{Amino acid embeddings visualized from ReLSO recapitulate their underlying biochemical properties in their organization. We compare the embeddings of the AE model $\alpha=0.0$ and the JT-AE ($\alpha=1.0$) model and observe that the joint-training approach produces better organized amino acid embeddings across the datasets. }
    \label{fig: aa-embeds}
\end{figure}








